\documentclass[final,3p,times,twocolumn]{elsarticle}
\usepackage{url}
\usepackage{verbatim}
\usepackage{graphicx}
\usepackage{cite}
\usepackage{times}
\usepackage{epsfig}
\usepackage{graphicx}
\usepackage{amssymb}
\usepackage{fontawesome}
\usepackage{amsmath}
\usepackage{mathabx}

\begin{document}

\begin{frontmatter}

\title{Enhancing Image Restoration through Learning Context-Rich and Detail-Accurate Features}


\author{Hu Gao$^a$, Depeng Dang$^a$$^*$}
\ead{gao_h@mail.bnu.edu.cn, ddepeng@bnu.edu.cn}
\affiliation{organization={Beijing Normal University, Artificial Intelligence},
            city={Beijing},
            postcode={100875}, 
            country={China}}

\begin{abstract}
Image restoration involves recovering high-quality images from their corrupted versions, requiring a nuanced balance between spatial details and contextual information. While certain methods address this balance, they predominantly emphasize spatial aspects, neglecting frequency variation comprehension. 
In this paper, we present a multi-scale design that optimally balances these competing objectives, seamlessly integrating spatial and frequency domain knowledge to selectively recover the most informative information.
Specifically, we develop a hybrid scale frequency selection block (HSFSBlock), which not only captures multi-scale information from the spatial domain, but also selects the most informative components for image restoration in the frequency domain. Furthermore, to mitigate the inherent noise introduced by skip connections employing only addition or concatenation, we introduce a skip connection attention mechanism (SCAM) to selectively determines the information that should propagate through skip connections. The resulting tightly interlinked architecture, named as LCDNet. Extensive experiments conducted across diverse image restoration tasks showcase that our model attains performance levels that are either superior or comparable to those of state-of-the-art algorithms. The code and the pre-trained models are released at \url{https://github.com/Tombs98/LCDNet}.
\end{abstract}

\begin{keyword}
Image restoration, multi-scale, skip feature fusion, frequency selection.
\end{keyword}

\end{frontmatter}
\section{Introduction}

Image restoration represents a longstanding low-level vision challenge aimed at recovering a latent sharp image from its corrupted version. This problem is often considered an inherently ill-posed inverse problem. Many conventional approaches~\citep{yang2020single, karaali2017edge}, tackle this issue by explicitly integrating various priors or handcrafted features to constrain the solution space to natural images. However, the design of such priors poses challenges and tends to lack generalizability.

In recent times, there has been significant progress in the development of deep neural networks for image restoration. Utilizing large-scale datasets, advanced models such as Convolutional Neural Networks (CNNs) and Transformers can implicitly learn more general priors by capturing intricate image features, resulting in superior performance compared to traditional methods. Over the years, a multitude of network modules and functional units for image restoration have been devised, either through the creation of new ones or the incorporation of advanced modules from other domains. This has given rise to a diverse range of models, including encoder-decoder architectures~\citep{neuCUI2024429,neuZHANG2025107378,gao2024learning}, multi-stage networks~\citep{Zamir2021MPRNet,gao2025mixed}, dual networks~\citep{PRWU2024110291,prgonzalez2024dgd}, recursive residual structures~\citep{ms9919385}, generative models~\citep{deganv2}, and various transformer enhancements~\citep{u2former,Zamir2021Restormer}.

Nevertheless, the methods mentioned above predominantly focus on the spatial domain, often neglecting the consideration of frequency variation knowledge between sharp and degraded image pairs. Although a limited number of methods~\citep{neuLIU2025106834,MRDNetzhang2024image,neuLIU2025107028,eswaxie2024mwa} leverage transformation tools to explore the frequency domain, these approaches primarily emphasize capturing global information and fall short in addressing how to effectively choose the most informative frequency for recovery. While efforts like~\citep{cui2025adair,gao2025frequency} either highlight or attenuate resulting frequency components to select the most informative frequency, they often grapple with finding a balance between spatial details and contextualized information.
\begin{figure*}
    \centering
    \includegraphics[width=1\linewidth]{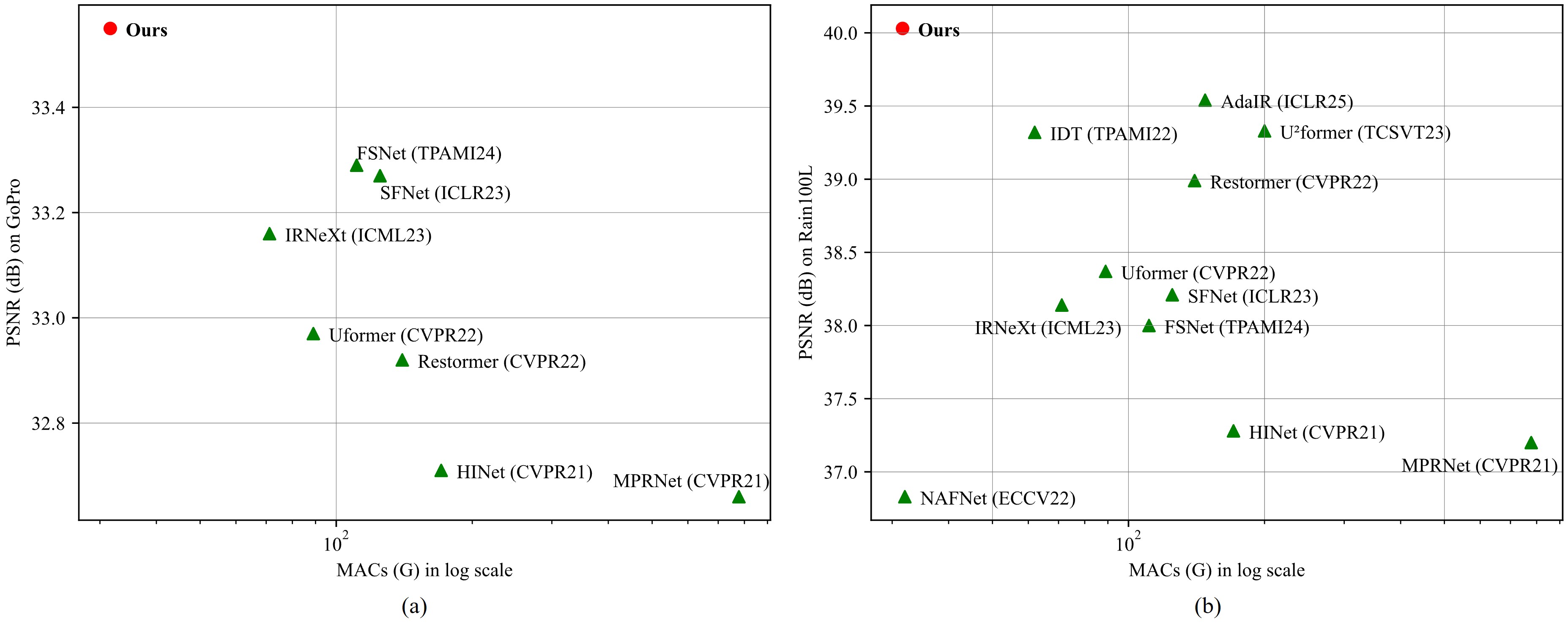}
    \caption{Computational cost vs. PSNR of models on Image Deblurring (left) and Image Deraining (right) tasks. Our LCDNet achieve the SOTA performance with up to 84.2\% of cost reduction on Image Deraining. MACs are computed on patch size of 256 × 256.}
    \label{fig:param}
\end{figure*}

Given the information provided, a pertinent question arises: Can we devise a network capable of learning context-rich and detail-accurate features for image restoration? To address this goal, we put forth a multi-scale design incorporating several key components.
In particular, we introduce a hybrid scale frequency selection block (HSFSBlock) designed to initially capture multi-scale information from the spatial domain by multi-scale spatial feature block (MSSFBlock). Subsequently, these context-rich spatial features are fed into the multi-branch  selective frequency module (MSFM). MSF generate high and low-frequency maps  using a learnable filter and select the most informative frequency component to reconstruct via the attention.

Furthermore, to address the inherent noise introduced by skip connections using only addition or concatenation, we introduce a skip connection attention mechanism (SCAM). SCAM utilizes an attention mechanism to discriminatively determine the information that should propagate through skip connections. Crucially, SCAM combines features from different scales while preserving their unique complementary properties. This enables our model to acquire a rich and detail-accurate set of features.

Finally, we adopt a coarse-to-fine mode to address challenges in model training, involving multiple scales of input and output. As illustrated in Figure~\ref{fig:param}, our model achieves state-of-the-art performance while maintaining computational efficiency compared to existing methods.

The main contributions of this work are:
\begin{enumerate}
	\item We propose LCDNet, a multi-scale design aimed at learning both spatial details and contextualized information for image restoration. Extensive experiments showcase that LCDNet achieves state-of-the-art performance across various image restoration tasks.
    \item We design a hybrid scale frequency selection block (HSFSBlock) which not only captures multi-scale information from the spatial domain but also selectively chooses the most informative components for image restoration through attention in the frequency domain. 
    \item We develop a skip connection attention mechanism (SCAM) to discriminatively determine the information that should propagate in skip connections, addressing noise introduced by conventional addition or concatenation.
\end{enumerate}

\section{Related Work}
\subsection{Image Restoration}
Image restoration aims to enhance the quality of an image by addressing various degradations. Given its inherently ill-posed nature, many traditional methods~\citep{yang2020single, karaali2017edge} tackle this challenge by incorporating hand-crafted priors to constrain the set of feasible solutions, often leading to a lack of generalizability.

With the rapid advancement of deep learning, numerous works based on this paradigm have gained significant popularity in the field of image restoration, showcasing more favorable performance compared to conventional methods.
MPRNet~\citep{Zamir2021MPRNet} decomposes the overall recovery process into manageable steps, striking a delicate balance between preserving spatial details and incorporating high-level contextualized information in image restoration.
MIRNet-V2~\citep{Zamir2022MIRNetv2} introduces a multi-scale architecture that maintains spatially-precise high-resolution representations throughout the network, gathering complementary contextual information from low-resolution representations.
NAFNet~\citep{chen2022simple} presents a streamlined baseline network for image restoration, achieved by either removing or replacing nonlinear activation functions.
DGD-cGAN~\citep{prgonzalez2024dgd} introduces a conditional generative adversarial network with dual generators to remove the colour cast of underwater images and restore their true colours.
Despite their successes, these methods predominantly rely on Convolutional Neural Networks (CNNs). The intrinsic properties of convolutional operations, constrain the models' capability to efficiently eliminate long-range degradation perturbations.

To address limitations, Transformers~\citep{2017Attention} have been employed in image restoration, demonstrating superior performance compared to previous CNN-based baselines. To enhance efficiency, Uformer~\citep{Wang_2022_CVPR}, SwinIR~\citep{liang2021swinir}, and U$^2$former~\citep{u2former} compute self-attention based on a window. Restormer~\citep{Zamir2021Restormer} introduces transposed attention, implementing self-attention across channels rather than the spatial dimension, resulting in an efficient transformer model. IDT~\citep{IDT} designs a window-based and spatial-based dual Transformer for excellent results. Despite their successes, these methods predominantly focus on the spatial domain, often neglecting consideration of frequency variation knowledge between sharp and degraded image pairs.

\subsection{Frequency Learning}

In recent years, there has been a trend in deep learning-based vision research to explore the frequency domain, driven by the advantages in handling different frequency subbands. For instance,~\citep{Chi_2020_FFC} introduces a generic fast Fourier convolution (FFC) operator to overcome the limitations of convolutional local receptive fields, achieving notable performance in high-level vision tasks.
Leveraging the convolution theorem, which equates convolution in the spatial domain to point-wise multiplication in the frequency domain, FFTformer~\citep{kong2023efficient} develops an effective and efficient method that harnesses the properties of Transformers for high-quality image deblurring.
UHDFour~\citep{fLi2023ICLR} incorporates Fourier transform into the network, processing the amplitude and phase of a low-light image separately to prevent noise amplification during luminance enhancement.
DeepRFT~\citep{fxint2023freqsel} integrates Fourier transform to incorporate kernel-level information into image deblurring networks.
In the quest to choose the most informative frequency component for reconstruction, AirFormer~\citep{10196308} introduces a supplementary prior module to selectively filter high-frequency components, thereby ensuring the preservation of low-frequency components. Similarly, SFNet~\citep{SFNet} and FSNet~\citep{FSNet} leverage a multi-branch and content-aware module to dynamically and locally decompose features into separate frequency subbands, enabling the selection of the most informative components for recovery.
However, these methods rely on simple addition or concatenation in the encoder and decoder skip connections, overlooking the potential impact of introducing implicit noise.

In this paper, we present  a hybrid scale frequency selection block (HSFSBlock) that captures multi-scale information from the spatial domain and selectively chooses the most informative frequency components. Additionally, a skip connection attention mechanism (SCAM) is developed to discern the information propagated in skip connections, mitigating noise introduced by conventional addition or concatenation.

\begin{figure*}[htb] 
	\centering
	\includegraphics[width=1\textwidth]{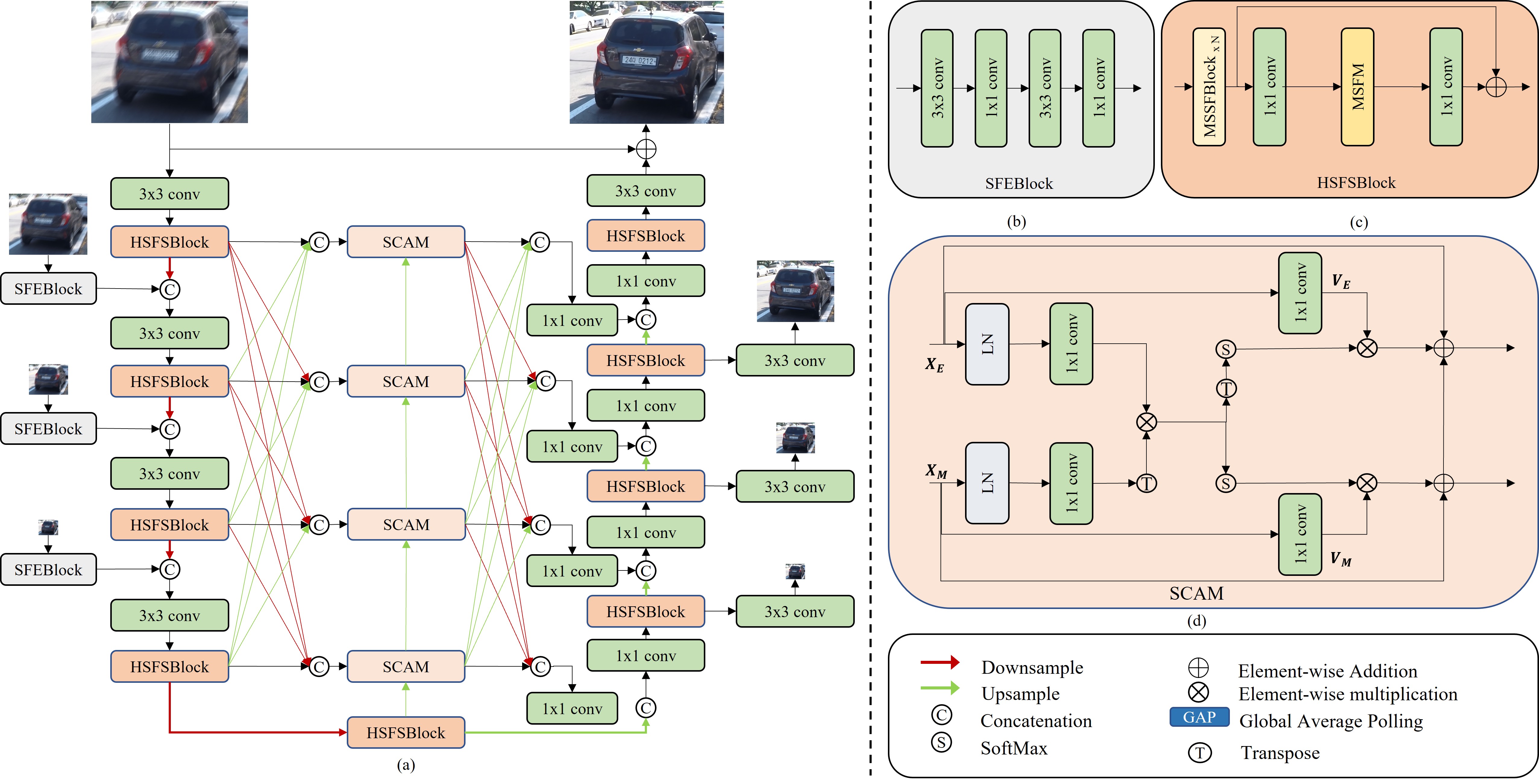}
	\caption{
(a) Overall architecture of LCDNet. (b) Shallow feature extraction block (SFEBlock) for capturing shallow features in low-resolution input images. (c) Hybrid scale frequency selection block (HSFSBlock) that captures multi-scale spatial information from the spatial domain and selects the most informative frequency components for image restoration. (d) Skip connection attention mechanism (SCAM) for discriminative information propagation in skip-connections.}
	\label{fig:network}
\end{figure*}

\section{Method}
In this section, we first outline the overall pipeline of our LCDNet. Subsequently, we delve into the details of the proposed hybrid scale frequency selection block (HSFSBlock), and skip connection attention mechanism (SCAM).

\subsection{Overall Pipeline} 
The overall pipeline of our proposed LCDNet, shown in Figure~\ref{fig:network}(a), is based on a  hierarchical encoder-decoder. 

Given a degraded image $\mathbf{I} \in \mathbb R^{H \times W \times 3}$, 
LCDNet initially applies a convolution to acquire shallow features $\mathbf{F_{0}} \in \mathbb R^{H \times W \times C}$  ($H, W, C$ are the feature map height, width, and channel number, respectively). 
These shallow features undergo a four-scale encoder to obtain in-depth features. 
In order to enhance training, we utilize multi-input and multi-output mechanisms.  Low-resolution degraded images are integrated into the main path through SFEBlock (see Figure~\ref{fig:network}(b)) and concatenation, followed by a 3 × 3 convolution to adjust channels. The in-depth features proceed to a middle block, and the resulting deepest features feed into a four-scale decoder, progressively restoring features to the original size. Throughout this process, encoder features, discerned through SCAM discrimination, are concatenated with decoder features to aid in the reconstruction.
Finally, we apply  convolution to the refined features to generate residual image $\mathbf{X}\in \mathbb R^{H \times W \times 3}$ to which degraded image is added to obtain the restored image: $\mathbf{\hat{I}} = \mathbf{X} +\mathbf{I}$.  It's important to note that the three low-resolution results are solely used for training. To facilitate the frequency selection process, we optimize the proposed network adopt $L_1$ loss in both spatial and frequency domains: 
\begin{equation}
\label{eq:loss}
\begin{aligned}
     L_{s} &= \sum_{i=1}^{4} \frac{||\mathbf{\hat{I_i}}-\mathbf{\overline I_i}||_1}{4}
   \\
    L_{f} &= \sum_{i=1}^{4} \frac{||\mathcal{F}(\mathbf{\hat{I}_i})-\mathcal{F}(\mathbf{\overline I_i})||_1}{4}
    \\
    L &= L_s + \lambda L_f 
\end{aligned}
\end{equation}
where $i$ denotes the index of input/output images at different scales, $\mathcal{F}$ represents fast Fourier transform, $\mathbf{\overline I_i}$ denotes the target images, and the parameter $\lambda$ controls the relative importance of the two loss terms, set to 0.1 as in~\citep{FSNet}.

\subsection{Hybrid Scale Frequency Selection Block }
Many image restoration methods have achieved success, primarily concentrating on the spatial domain and neglecting frequency variation understanding. While some approaches utilize transformation tools to explore the frequency domain, they often lack the flexibility to selectively choose the most informative frequencies. To overcome this limitation, we introduce the hybrid scale frequency selection block (HSFSBlock). This block not only captures multi-scale information from the spatial domain but also selectively chooses the most informative frequency components for image restoration. Formally, given the input features at $(l-1)$ -th level $X_{l-1}$, the procedures of HSFSBlock can be defined as:
\begin{equation}
\begin{aligned}
\label{eq:hsfs1}
   X^{'}_{l} &= MSSFBlock_N(... MSSFBlock_1(X_{l-1} ...)
   \\
   X_{l} &= X^{'}_{l} \oplus f_{1 \times 1}^c(MSFM(f_{1 \times 1}^c(X^{'}_{l})))
\end{aligned}
\end{equation}
where  $f_{1 \times 1}^c$ represents $1 \times 1$ convolution; $ X^{'}_{l}$ and $ X_{l}$ denote the output from $N$ multi-scale spatial feature blocks (MSSFBlock) and multi-branch selective frequency module (MSFM). 

\subsubsection{Multi Scale Spatial Feature Block}

\begin{figure}[htb] 
	\centering
	\includegraphics[width=0.5\textwidth]{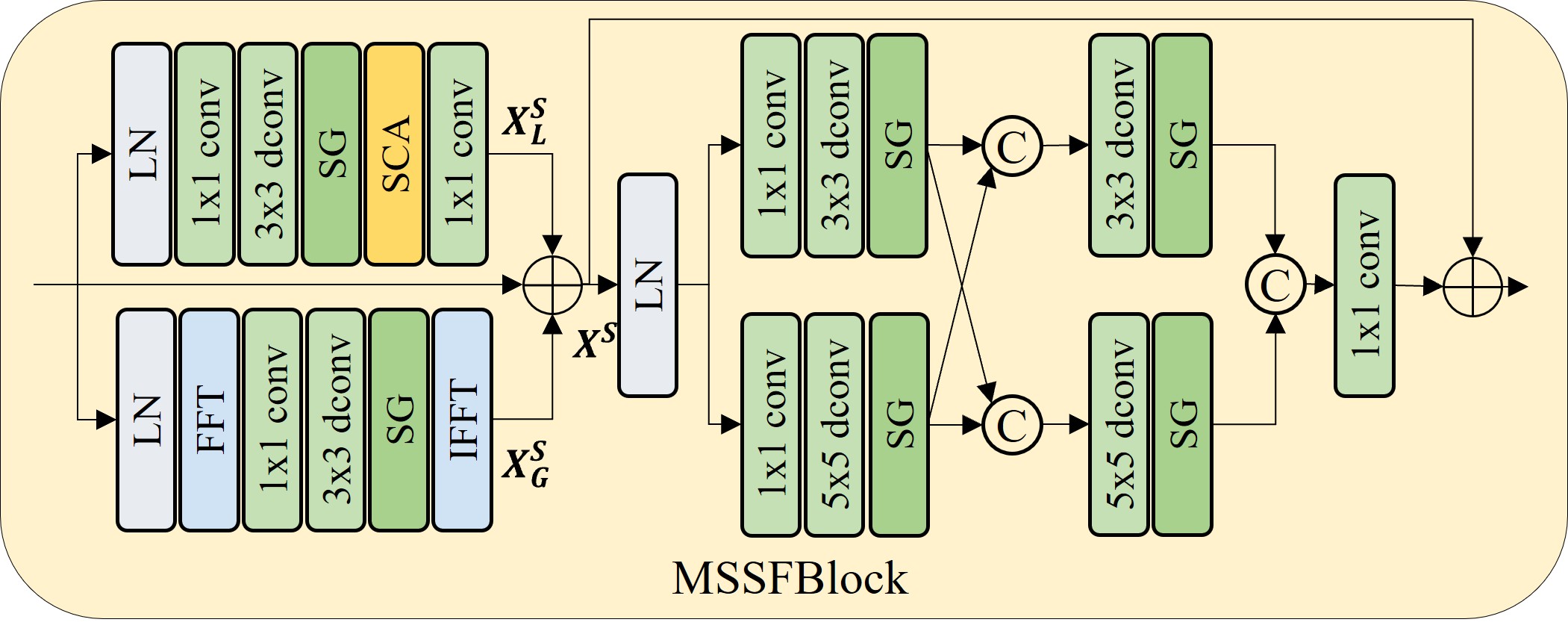}
	\caption{The structure of MSSFBlock. }
	\label{fig:nafnet}
\end{figure}

The effectiveness of rich multi-scale representations has been fully demonstrated~\citep{Zamir2020MIRNet, Zamir2022MIRNetv2} in improving image restoration performance. In this context, we  propose a multi scale block spatial feature block (MSSFBlock) designed to simultaneously preserve precise spatial details and acquire contextually enriched feature representations across multiple spatial scales.

As shown in Figure~\ref{fig:nafnet}, the MSSFBlock first processes the original high-resolution features using a two-branch structure—comprising a local branch (top) and a global branch (bottom)—to preserve detailed information. It then incorporates two multi-scale depth-wise convolutional paths to capture contextual features at multiple scales. Specifically, given an input tensor $X \in \mathbb R^{H \times W \times C}$, we initially process it at the original resolution to obtain spatially detailed features $X^{s}$ as follows:
\begin{equation}
\begin{aligned}
	\label{equ:0msb}
	X^{S}_{L} &= f_{1 \times 1}^c(SCA(SG(f_{3 \times 3}^{dwc} (f_{1 \times 1}^c(LN(X)))))
\\
	X^{S}_{G} &= \mathcal{IF}(SG(f_{3 \times 3}^{dwc} (f_{1 \times 1}^c(\mathcal{F}(LN(X)))))
\\
    X^{S} &= X^{S}_{L} \oplus X^{S}_{G} \oplus X
\end{aligned}
\end{equation}
where  $f_{3 \times 3}^{dwc}$ denotes the $3 \times 3$ depth-wise convolution, $ \mathcal{IF}$ represents the inverse operation of the fast Fourier transform. $SG(\cdot)$~\citep{chen2022simple} begins by splitting a feature into two features along the channel dimension. Following this, $SG$ computes these features using a linear gate. SCA~\citep{chen2022simple} is employed as a simplified version of channel attention, calculate as follows

\begin{equation}
	\label{equ:01msb}
    SCA(X_{f}) = X_{f} \otimes f_{1 \times 1}^c( GAP(X_{f}))
\end{equation}

Subsequently, we input the spatial detail features $X^{s}$ into two multi-scale depth-wise convolutional paths to obtain context-rich features $X^{C}$. The entire procedure can be formulated as:
\begin{equation}
\begin{aligned}
	\label{equ:1msb}
	X^{t1} &= SG(f_{3 \times 3}^{dwc} (f_{1 \times 1}^c (LN(X^{S}))))
 \\
 X^{b1} &= SG(f_{5 \times 5}^{dwc} (f_{1 \times 1}^c (LN(X^{S}))))
 \\
 X^{t2} &= SG(f_{3 \times 3}^{dwc}([X^{t1}, X^{b1}]))
 \\
 X^{b2} &= SG(f_{5 \times 5}^{dwc}([X^{b1}, X^{t1}]))
 \\
 X^{C} &=  f_{1 \times 1}^c([X^{t2}, X^{b2}])
\end{aligned}
\end{equation}
where $[\cdot]$ represents the channel-wise concatenation. Then we obtain precise spatial details and contextually enriched feature representations $X^{SC}$ as follows:
\begin{equation}
    \label{equ:2msb}
    X^{SC} = X^{S} \oplus X^{C}
\end{equation}

Finally, we input the feature $X^{SC}$ in the spatial domain to the MSFM for frequency selection and choose the most informative component for image restoration.

\begin{figure*}[htb] 
	\centering
	\includegraphics[width=1\textwidth]{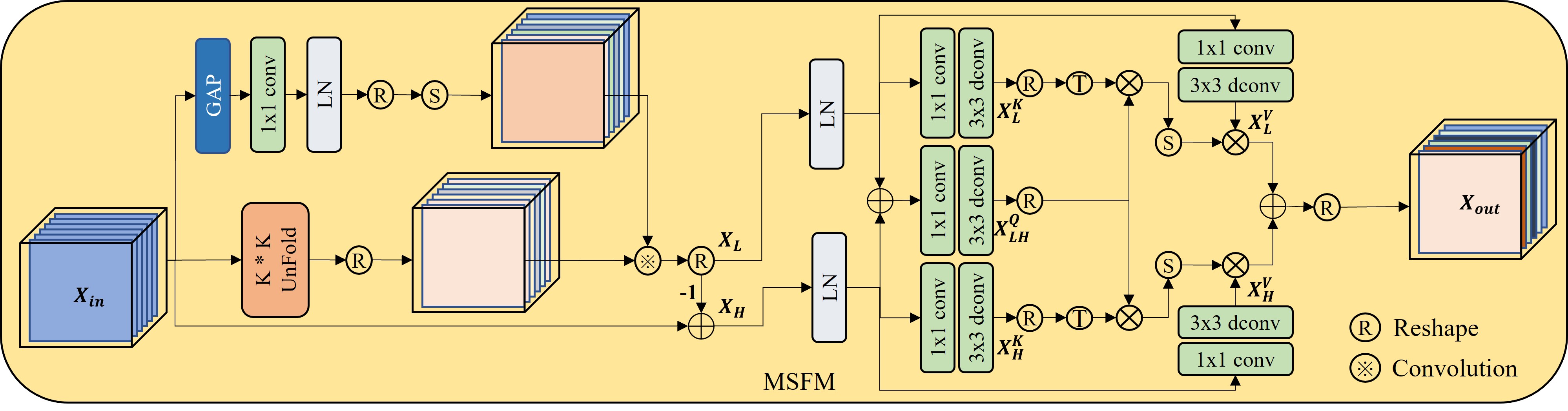}
	\caption{The structure of  multi-branch selective frequency module (MSFM). }
	\label{fig:dfs}
\end{figure*}

\subsubsection{Multi-branch Selective Frequency Module }
To facilitate the automatic selection of the most informative components for recovery, we draw inspiration from previous work~\citep{FSNet} and propose the multi-branch selective frequency module (MSFM). Illustrated in Figure~\ref{fig:dfs}, MSFM initially generates high and low-frequency features dynamically using a learnable filter, followed by discriminating which low-frequency and high-frequency information to retain. We only show the one-branch case for simplicity.

To elaborate, with an input feature $X_{in} \in \mathbb{R}^{H \times W \times C}$, we initially employ Global Average Pooling (GAP), convolution, Layer Normalization (LN), reshaping, and a softmax activation function to generate a low-pass filter for each group of the input, as outlined below:
\begin{equation}
\label{eq:fl}
F_l = Softmax(R(LN( f_{1 \times 1}^c(GAP(X_{in} )))))
\end{equation}
where $R(\cdot)$ denotes a reshape operation that divides the feature map from $\mathbb R^{1 \times 1 \times gk^2}$ to
$\mathbb R^{g \times k \times k}$, $g$ denotes the number
of groups, $k \times k$ is the kernel size of low-pass filters. We utilize $k \times k$ kernel sizes to unfold the input feature $X_{in}$ and reshape it. The resulting output is convolved with the low-pass filter $F_l$, and subsequently reshaped back to the original feature map size to obtain the final low-frequency feature map $X_L$. This process can be expressed as:
\begin{equation}
	\label{equ:fl}
	X_L =  F_l \ocoasterisk R(UnFold(X_{in}))
\end{equation}

To obtain the high-frequency feature map $X_H$, we subtract the resulting low-frequency feature map $X_L$  from the input feature $X_{in}$, which is expressed as:
\begin{equation}
	\label{equ:fh}
	X_H =  X_{in} - X_L
\end{equation}

Subsequently, we determine which low-frequency and high-frequency information should be retained through attention mechanisms.  Specifically, we first perform layer normalization on $X_L$ and $X_H$, and then aggregate the channel context information by $1 x 1$ and $3 x 3 $ convolution to generate the matrix $Q, K, V$ for computing attention. It is worth noting that low-frequency and high-frequency information are first fused by sum before generating the query matrix $Q$. The overall process is as follows:

\begin{equation}
\begin{aligned}
\label{equ:sff}
\overline{X_L}, \overline{X_H} &= LN(X_L), LN(X_H)
\\
X^K_L & = f_{3 \times 3}^{dwc}(f_{1 \times 1}^c(\overline{X_L}))
\\
X^K_H & = f_{3 \times 3}^{dwc}(f_{1 \times 1}^c(\overline{X_H}))
\\
X^V_L & = f_{3 \times 3}^{dwc}(f_{1 \times 1}^c(\overline{X_L}))
\\
X^V_H & = f_{3 \times 3}^{dwc}(f_{1 \times 1}^c(\overline{X_H}))
\\
X^Q_{LH} & = f_{3 \times 3}^{dwc}(f_{1 \times 1}^c(\overline{X_L} \oplus \overline{X_H}))
\end{aligned}
\end{equation}

Finally,  we compute attention discriminative between low-frequency and high-frequency features to retain the most informative components and effectively fuse for reconstruction as follows:
\begin{equation}
\begin{aligned}
\label{equ:sff2}
   F_L &= Sofamax((R(X^K_L))^T \otimes R(X^Q_{LH})) \otimes X^V_L
   \\
    F_H &= Sofamax((R(X^K_H))^T \otimes R(X^Q_{LH})) \otimes X^V_H
    \\
    X_{out} &= R(F_L \oplus F_H)
\end{aligned}
\end{equation}
where the reshape operation enables self-attention to be applied across channels instead of the spatial dimension, thereby reducing both time and memory complexity.

\subsection{Skip Connection Attention Mechanism (SCAM)}
We observe that features from the encoder may contain image degradation factors, and the simple addition or concatenation process for skip connections between the encoder and decoder is susceptible to implicit noise, potentially impacting the model's image restoration capability.

Based on this observation,  we introduce a skip connection attention mechanism (SCAM) to selectively determines the information that should propagate through skip connections. Specifically, for the SCAM at the lowest level, we initially concatenate the  encoder features $E_i, (i=1,2,3,4)$  and adjust the number of channels through a convolution:
\begin{equation}
\label{eq:SCAM1}
        X_{E1} = f_{1 \times 1}^c([E_1, E_2, E_3, E_4])
\end{equation}
Subsequently, we obtain $X_{M1}$ by upsampling the feature map of the middle block, and  apply layer normalization. 
As illustrated in Figure~\ref{fig:network} (d), then we compute the dot products between the query $Q_{E1}, Q_{M1}\in \mathbb R^{H \times W \times C}$ projected by the  $X_{E1}$ and the $X_{M1}$. Noted that, $K_{E1} = Q_{M1}^{'}$, $K_{M1} = Q_{E1}$. Next, we generate the value matrix $V_{E1}$ and $V_{M1}$ by using a $1 \times 1$ convolution layer, respectively. 
Subsequently, we compute bidirectional cross-attention  as follows:
\begin{equation}
    \begin{aligned}
        F_{M1 \rightarrow E1} &= Attention(Q_{M1}, K_{M1} , V_{E1})
        \\
        F_{E1 \rightarrow M1} &= Attention(Q_{E1}, K_{E1}, V_{M1})
    \end{aligned}
\end{equation}
where
\begin{equation}
	\label{equ:102}
    Attention(Q, K, V) =  Softmax(\frac{QK^T}{\beta})V
\end{equation}
where $\beta$ is a learning scaling parameter used to adjust the magnitude of the dot product of $Q$ and $K$ prior to the application of the Softmax function defined by $\beta = \sqrt{C}$. We fuse the interacted information $ F_{M1 \rightarrow E1}$ and $ F_{E1 \rightarrow M1}$ by element-wise addition to obtain the most useful information $\hat{F}_{E1}$, 
\begin{equation}
\label{eq:rcam}
        \hat{F}_{E1}= \lambda_L F_{M1 \rightarrow E1} \oplus \lambda_H F_{E1 \rightarrow M1}
\end{equation}
where $\lambda_L$ and $\lambda_H$ are channel-wise scale parameters that are trainable and initialized with zeros to aid in stabilizing training. 

Through the aforementioned process, we obtain the encoder features $\hat{F}_{Ei}, (i=1,2,3,4)$ that retain useful information for all levels. Finally, the features ($\hat{F}_{Ei}$) are concatenated, and a convolution operation is applied to adjust the number of channels, resulting in the final feature that will fuse with the decoder. SCAM provides two key advantages. Firstly, it selectively identifies the information to be transmitted in the skip connection, mitigating the interference of implicit noise. Secondly, it combines features at different scales while preserving their distinctive complementary properties. 

\begin{figure*}[htb] 
	\centering
	\includegraphics[width=1\textwidth]{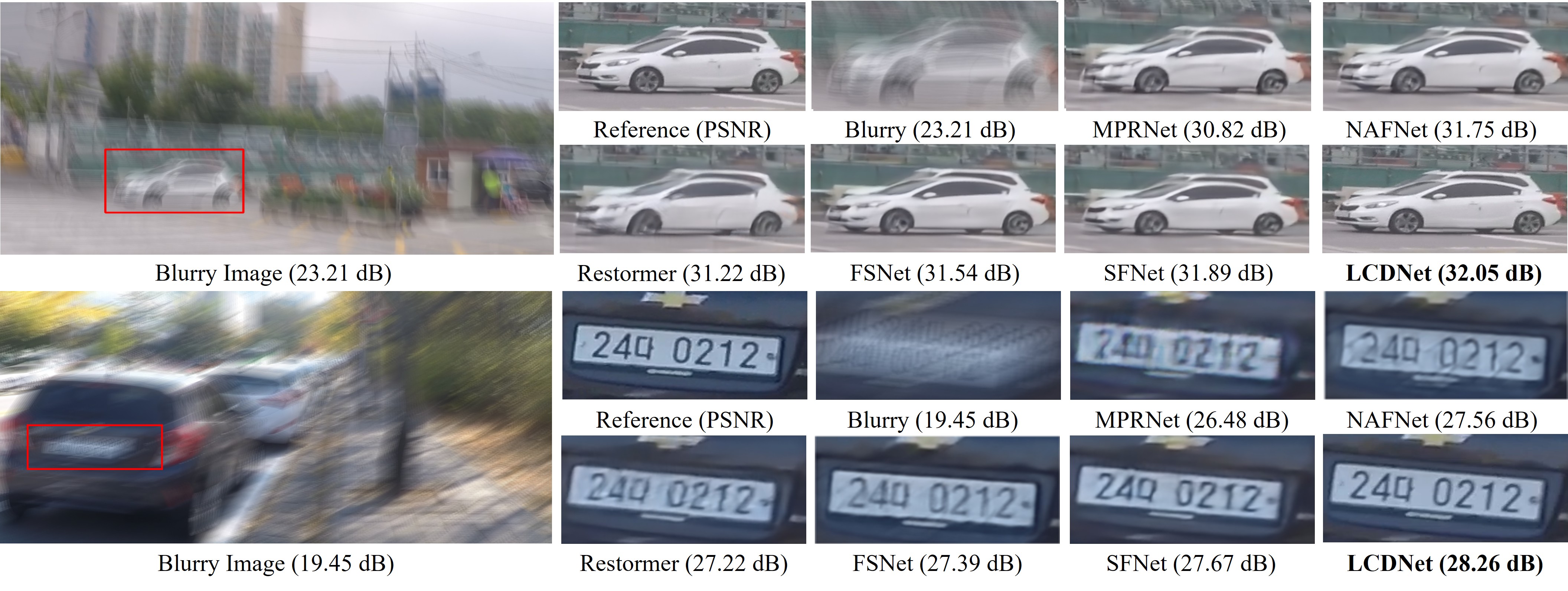}
	\caption{\textbf{Image motion deblurring} comparisons on the GoPro dataset~\citep{Gopro}. Compared to the state-of-the-art methods, our LCDNet excels in restoring sharper and perceptually faithful images. }
	\label{fig:blurm}
\end{figure*}

\section{Experiments}
In this section, we offer details about the experimental settings and subsequently present both qualitative and quantitative comparisons between LCDNet and other state-of-the-art methods. Following that, we conduct ablation studies to validate the effectiveness of our approach. Finally, we assess the resource efficiency of LCDNet.

\subsection{Experimental Settings}
In this section, we introduce the details of the used datasets, and training details.

\subsubsection{\textbf{Datasets}}

Adhering to the experimental setups of recent state-of-the-art methods~\citep{FSNet,Zamir2021MPRNet}, for the image motion deblurring task, LCDNet is trained using the GoPro dataset\citep{Gopro}. This dataset comprises 2,103 image pairs for training and 1,111 pairs for evaluation. To assess the generalizability of our approach, the GoPro-trained model is directly applied to the test images of the HIDE dataset~\citep{HIDE}, encompassing 2,025 images for evaluation. Both the GoPro and HIDE datasets are synthetically generated, but the RealBlur~\citep{realblurrim_2020_ECCV} dataset comprises image pairs captured under real-world conditions. This dataset includes two subsets: RealBlur-J,  and RealBlur-R. For single-image defocus deblurring,  we utilize the DPDD dataset~\citep{DPDNet}, which comprises images from 500 indoor/outdoor scenes. LCDNet is trained using the center view images as input, computing loss values between outputs and corresponding ground-truth images. For the image deraining task, our model is trained using 13,712 clean-rain image pairs collected from various datasets~\citep{Rain100,Test100,8099669,7780668}. Using the trained LCDNet, evaluations are conducted on different test sets, including Rain100H~\citep{Rain100}, Rain100L~\citep{Rain100}, Test100~\citep{Test100}, and Test1200~\citep{DIDMDN}.

\subsubsection{\textbf{Training details}}
For different tasks, separate models are trained, and unless specified otherwise, the following parameters are utilized. The models are trained using the Adam optimizer~\citep{2014Adam} with parameters $\beta_1=0.9$ and $\beta_2=0.999$. The initial learning rate is set to $2 \times 10^{-4}$ and gradually reduced to $1 \times 10^{-7}$ using the cosine annealing strategy~\citep{2016SGDR}. The batch size is chosen as $32$, and patches of size $256 \times 256$ are extracted from training images. Data augmentation includes horizontal and vertical flips.

\begin{figure*}[htb] 
	\centering
	\includegraphics[width=1\linewidth]{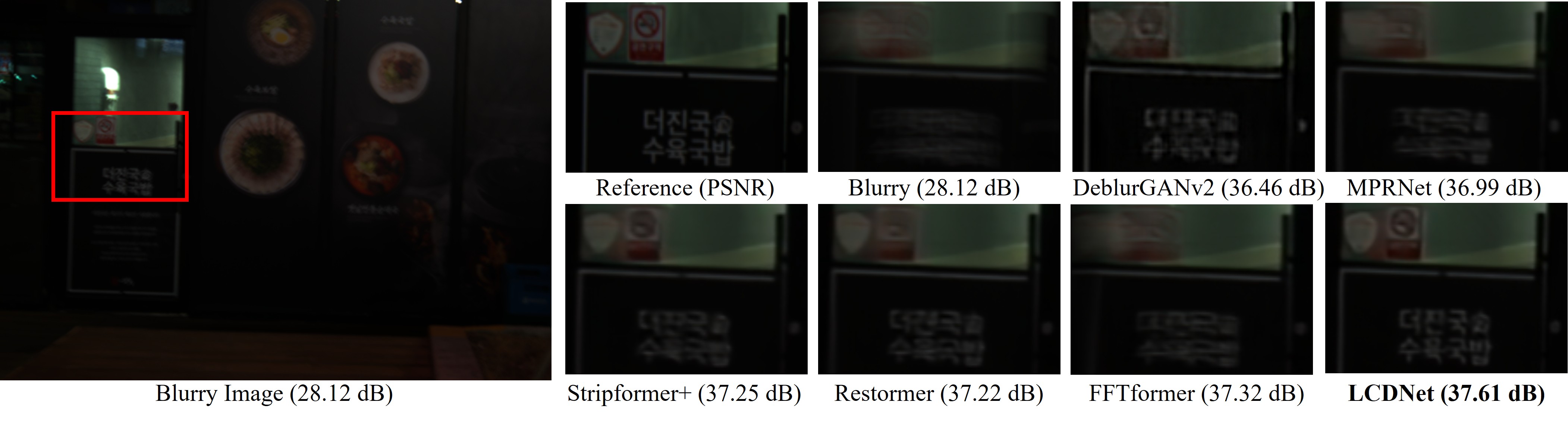}
	\caption{\textbf{Comparison of image motion deblurring} on the RealBlur dataset~\citep{realblurrim_2020_ECCV}.}
	\label{fig:real2}
\end{figure*}

\begin{table*}
\centering
\caption{Image motion deblurring results.  The best and second best scores are \textbf{highlighted} and \underline{underlined}. Our LCDNet  are trained only on the GoPro dataset~\citep{Gopro}but achieves the highest scores on the average of the effects on both datasets. \label{tb:deblur}}

\begin{tabular}{ccccc||cc}
    \hline
    \multicolumn{1}{c}{} & \multicolumn{2}{c}{GoPro~\citep{Gopro}}  & \multicolumn{2}{c||}{HIDE~\citep{HIDE}} & \multicolumn{2}{c}{Average}
    \\
   Methods & PSNR $\uparrow$ & SSIM $\uparrow$ & PSNR $\uparrow$ & SSIM $\uparrow$   &  PSNR $\uparrow$ &  SSIM $\uparrow$
    \\
    \hline\hline
    MPRNet~\citep{Zamir2021MPRNet} & 32.66 & 0.959 & 30.96 & 0.939 &31.81  &0.949
    \\
    Uformer~\citep{Wang_2022_CVPR} &32.97 & \textbf{0.967} &30.83 &\underline{0.952} &31.90  &\underline{0.960}
     \\
    MSFS-Net~\citep{MSFSnet} & 32.73 & 0.959 & 31.05 & 0.941& 31.99 & 0.950 
    \\
    Restormer~\citep{Zamir2021Restormer} & 32.92 & 0.961 & \underline{31.22} & 0.942 &32.07 &0.952
    \\
    IRNeXt~\citep{IRNeXt} &33.16 &0.962 &- & - &- &-
    \\
    SFNet~\citep{SFNet} &33.27 &0.963 &31.10 & 0.941 &\underline{32.19} &0.952
    \\
    MRDNet+~\citep{MRDNetzhang2024image}&32.47& 0.959 &30.54 &0.939&31.51&0.949
    \\
    FSNet~\citep{FSNet} &\underline{33.29}&0.963 &31.05 & 0.941 &32.17 &0.952
    \\
    \hline
    \textbf{LCDNet(Ours)} &\textbf{33.55}&\underline{0.966}&\textbf{31.51}&\textbf{0.958} &\textbf{32.53} &\textbf{0.962}
    \\
    \hline
\end{tabular}
\end{table*}

\subsection{Experimental Results}
\subsubsection{Image Motion Deblurring}
To ensure fair comparisons, we adhere to the protocols of this dataset and either retrain or fine-tune the deep learning methods that were not originally trained on this dataset. The quantitative results on the GoPro and HIDE datasets are presented in Table~\ref{tb:deblur}.
Our LCDNet attains the highest PSNR and SSIM values on the average effects for both datasets. Specifically, in comparison to the state-of-the-art CNN-based method FSNet~\citep{FSNet}, our method achieves a 0.26 dB higher PSNR gain than FSNet~\citep{FSNet}. When compared to the state-of-the-art Transformer-based method Restormer~\citep{Zamir2021Restormer}, our method achieves a 0.63 dB improvement.

Furthermore, as depicted in Figure~\ref{fig:param}(a), our LCDNet not only achieves superior performance but also incurs lower computational costs. It is noteworthy that, despite being trained exclusively on the GoPro~\citep{Gopro} dataset, our network still attains the best performance on the HIDE~\citep{HIDE} dataset, demonstrating its impressive generalization capability. Figure~\ref{fig:blurm} visually showcases that our model produces more aesthetically pleasing results.

\begin{figure*}[htb] 
	\centering
	\includegraphics[width=1\textwidth]{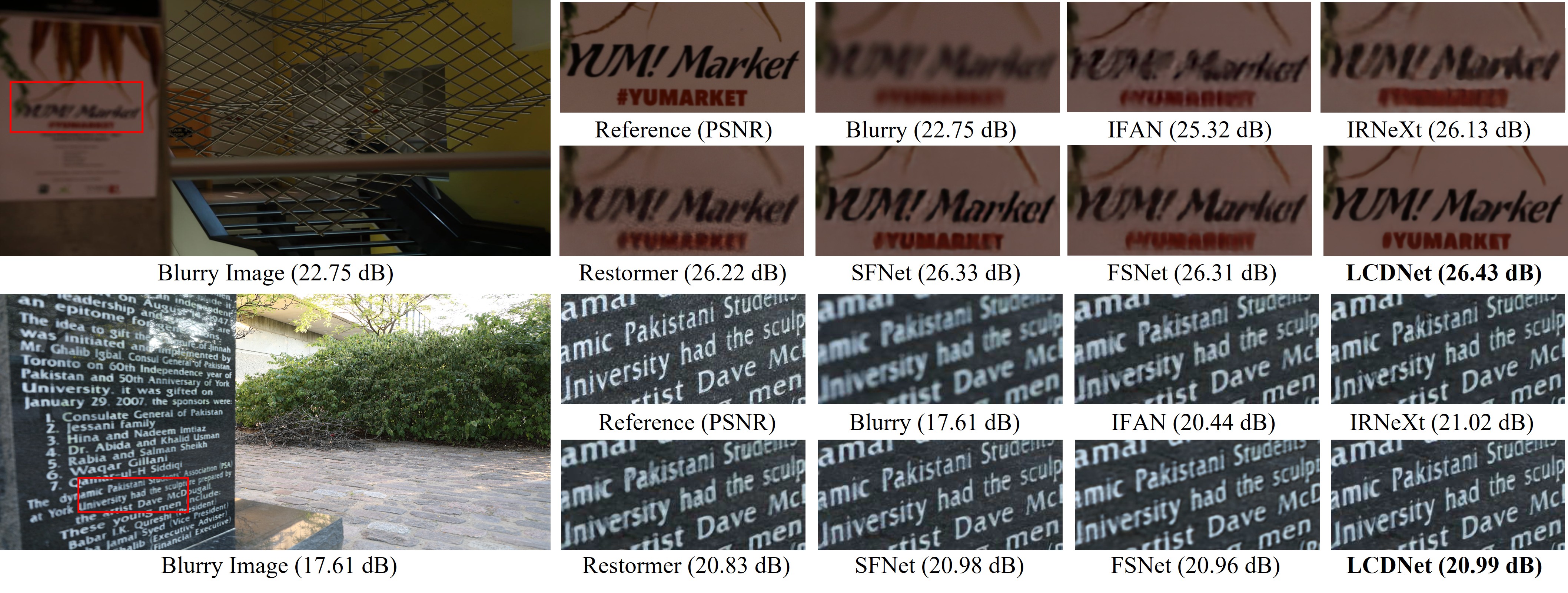}
	\caption{\textbf{Single image defocus deblurring} comparisons on the DDPD dataset~\citep{DPDNet}. Compared to the state-of-the-art methods, our LCDNet effectively removes blur while preserving the fine image details. }
	\label{fig:blurd}
\end{figure*}

\begin{figure*}[htb] 
	\centering
	\includegraphics[width=1\textwidth]{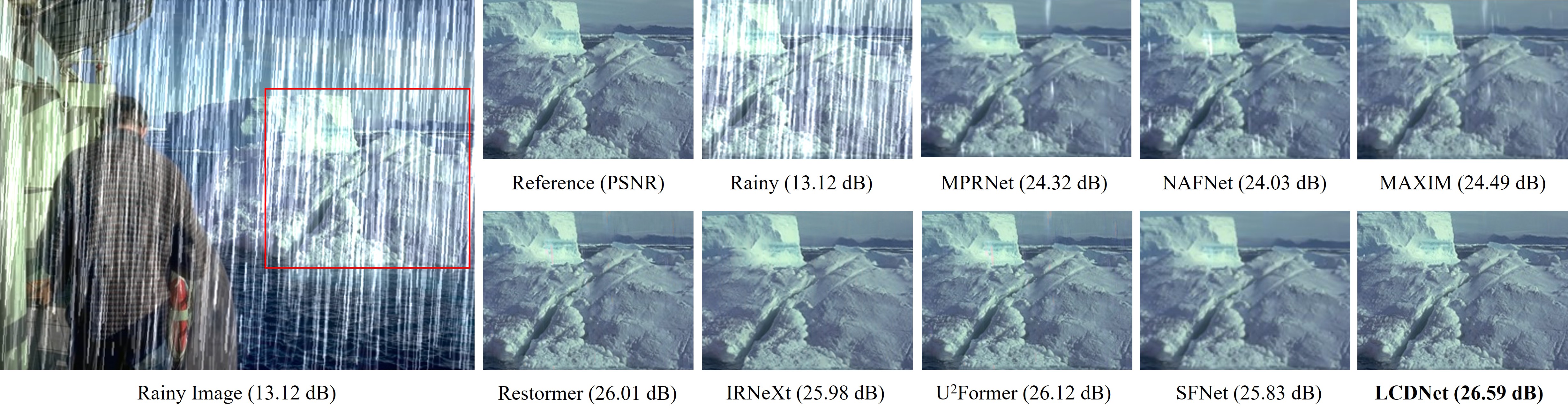}
	\caption{\textbf{Image  deraining} example on the Rain100H~\citep{Rain100}. 
The outputs of LCDNet show no traces of rain streaks. }
	\label{fig:rain}
\end{figure*}

\begin{table}
\centering
\caption{Quantitative real-world deblurring results.}
\label{tb:0deblurringreal}
 \resizebox{\linewidth}{!}{
\begin{tabular}{ccccc}
    \hline
    \multicolumn{1}{c}{} & \multicolumn{2}{c}{RealBlur-R}  & \multicolumn{2}{c}{RealBlur-J} 
    \\
   Methods & PSNR $\uparrow$ & SSIM $\uparrow$ & PSNR $\uparrow$ & SSIM $\uparrow$   
    \\
    \hline
DeblurGAN-v2~\citep{deganv2} & 36.44 & 0.935& 29.69& 0.870
\\
    MPRNet~\citep{Zamir2021MPRNet} & 39.31 & 0.972 & 31.76 & 0.922
   \\
Stripformer~\citep{Tsai2022Stripformer} & 39.84 & 0.975& 32.48 & 0.929
\\
FFTformer~\citep{kong2023efficient}&40.11& 0.973 &32.62 &0.932
\\
MHNet~\citep{gao2025mixed} &\underline{40.33} &\underline{0.976} &\underline{32.76} &\underline{0.933}
     \\
\hline
\textbf{LCDNet(Ours)}&\textbf{40.54}&\textbf{0.977}&\textbf{32.89}&\textbf{0.935}
    \\
    \hline
\end{tabular}}
\end{table}

We also evaluate our LCDNet on real-world images from the RealBlur dataset~\citep{realblurrim_2020_ECCV}. The results are presented in Table~\ref{tb:0deblurringreal}, our LCDNet achieves performance gains of 0.21 dB on the RealBlur-R subset over MHNet~\citep{gao2025mixed} and 0.13 dB on the RealBlur-J subset. Figure~\ref{fig:real2} presents visual comparisons of the evaluated approaches. Overall, the images restored by our model exhibit sharper details.

\begin{table*}
    \centering
        \caption{Quantitative comparisons with previous leading single-image defocus deblurring methods on the DPDD testset~\citep{DPDNet} (containing 37 indoor and 39 outdoor scenes).}
    \label{tab:deblurd}
    \resizebox{\linewidth}{!}{
    \begin{tabular}{c|ccc|ccc|ccc}
        \hline
    \multicolumn{1}{c|}{} & \multicolumn{3}{c|}{Indoor Scenes}  & \multicolumn{3}{c|}{Outdoor Scenes} & \multicolumn{3}{c}{Combined}
    \\
   Methods & PSNR $\uparrow$ & SSIM $\uparrow$ &MAE $\downarrow$  & PSNR $\uparrow$ & SSIM $\uparrow$  &MAE $\downarrow$    &  PSNR $\uparrow$ &  SSIM $\uparrow$ &MAE $\downarrow$  
    \\
    \hline
    \hline

Restormer~\citep{Zamir2021Restormer}& 28.87 &0.882 &0.025 &23.24 &0.743 &0.050  &25.98 &0.811 &0.038 
\\
IRNeXt~\citep{IRNeXt} &29.22 &0.879 &0.024  &23.53 &\underline{0.752} &\underline{0.049}  &26.30 &0.814&\underline{0.037} 
\\
SFNet~\citep{SFNet} &29.16 &0.878 &\underline{0.023}  &23.45 &0.747 &\underline{0.049}  &26.23 &0.811 &\underline{0.037} 
\\
FSNet~\citep{FSNet} &29.14 &0.878 &0.024  &23.45 &0.747 &0.050  &26.22 &0.811 &\underline{0.037} 
\\
 ALGNet~\citep{gao2024learning} &\underline{29.37} &\textbf{0.898} &\underline{0.023}  & \underline{23.68} &\textbf{0.755} &\textbf{0.048} & \underline{26.45} &\textbf{0.821} &\textbf{0.036 }
\\
\hline
\textbf{LCDNet(Ours)}&\textbf{29.45}	&\underline{0.892}	&\textbf{0.021}		&\textbf{23.69}	&\textbf{0.755}	&\textbf{0.048}		&\textbf{26.57}	&\underline{0.819}	&\textbf{0.036}	
    \\
    \hline
    \end{tabular}}
\end{table*}

\subsubsection{Single-Image Defocus Deblurring}
Table~\ref{tab:deblurd} presents a numerical comparison of defocus deblurring methods. LCDNet outperforms other state-of-the-art methods across all metrics. Notably, in the combined scenes category, LCDNet exhibits a 0.12 dB improvement over the leading method ALGNet~\citep{gao2024learning}.  The visual results in Figure~\ref{fig:blurd} illustrate that our method recovers more details and visually aligns more closely with the ground truth compared to other algorithms.

\begin{table*}
\centering
\caption{Image deraining results. When averaged across all four datasets, our LCDNet is better than the state-of-the-art by 0.61 dB. }\label{tb:derain}
\resizebox{\linewidth}{!}{
\begin{tabular}{ccccccccc||cc}
    \hline
    \multicolumn{1}{c}{} & \multicolumn{2}{c}{Test100~\citep{Test100}}  & \multicolumn{2}{c}{Test1200~\citep{DIDMDN}} & \multicolumn{2}{c}{Rain100H~\citep{Rain100}} & \multicolumn{2}{c||}{Rain100L~\citep{Rain100}} & \multicolumn{2}{c}{Average} 
    \\
   Methods &PSNR $\uparrow$ &  SSIM $\uparrow$  & PSNR $\uparrow$ & SSIM $\uparrow$ &PSNR $\uparrow$ &SSIM $\uparrow$ & PSNR $\uparrow$&SSIM $\uparrow$ &PSNR $\uparrow$ & SSIM $\uparrow$
    \\
    \hline\hline

     MPRNet~\citep{Zamir2021MPRNet}  & 30.27 & 0.907 & 32.91 &  0.916   & 30.51 & 0.890  & 37.20 & 0.965 & 32.73 & 0.921
       \\
   
     Restormer~\citep{Zamir2021Restormer} &\underline{32.00} & \textbf{0.923} & 33.19 & \underline{0.926} & 31.46 &0.904 &38.99 &0.978 &33.91 & \underline{0.933}
     \\

         U$^2$Former~\citep{u2former} & - & - & \underline{33.48} &\underline{0.926} & 30.87 & 0.893 & 39.31 & 0.982 & - &-
     \\
     MDARNet~\citep{MDARNet} & 28.98 & 0.892 & 33.08 & 0.919 & 29.71 & 0.884 & 35.68 & 0.961 & 31.86 &0.914
     \\
       IRNeXt~\citep{IRNeXt} &31.53 &0.919 &- & - &31.64 &0.902 & 38.14&0.972 &- &-
    \\
    SFNet~\citep{SFNet} &31.47&0.919&32.55&0.911 & \underline{31.90} & \textbf{0.908} & 38.21&0.974 &33.53&0.928
    \\
    FSNet~\citep{FSNet} &31.05&0.919 &33.08&0.916 & 31.77&\underline{0.906} &38.00 & 0.972 &33.48 &0.928
    \\
     MHNet~\citep{gao2025mixed} &31.25 &0.901 &33.45 &0.925 &31.08 &0.899 &\textbf{40.04} &\textbf{0.985} &\underline{33.96} &0.928
     \\
     \hline
      \textbf{LCDNet(Ours)}  & \textbf{32.27} & \underline{0.922} & \textbf{34.01}&\textbf{0.927} & \textbf{31.97}&\textbf{0.908} &\underline{40.01}&\underline{0.984} &\textbf{34.57}&\textbf{0.935}
    \\
    \hline
\end{tabular}}
\end{table*}

\subsubsection{Image Deraining}
Building on prior work~\citep{SFNet}, we compute PSNR/SSIM scores using the Y channel in the YCbCr color space for the image deraining task. Table~\ref{tb:derain} illustrates that our method consistently outperforms existing approaches across all four datasets. Notably, our method achieves a remarkable average improvement of 0.61 dB over all datasets compared to the best-performing method MHNet~\citep{gao2025mixed}. Furthermore, on the Test1200 dataset~\citep{DIDMDN}, LCDNet exhibits a substantial 0.53 dB PSNR improvement over the previous best method U$^2$former. In addition to quantitative assessments, Figure~\ref{fig:rain} presents qualitative results, showcasing the effectiveness of LCDNet in removing rain streaks of various orientations and magnitudes while preserving the structural content of the images.

\subsection{Ablation Studies}
The ablation studies are conducted on image motion deblurring (GoPro~\citep{Gopro}) to analyze the impact of each of our model components. Next, we describe the impact of each component.

\begin{table}
    \caption{Ablation studies of each module.}
    \label{tab:ablem}
    \centering
    \resizebox{\linewidth}{!}{
    \begin{tabular}{ccccc}
    \hline
       Multi-in, Multi-out & SCAM & HSFSBlock & PSNR & SSIM
       \\
       \hline
       \faTimes & \faTimes & \faTimes & 32.83 & 0.960
         \\
          \faCheck & \faTimes & \faTimes &32.92 & 0.960
          \\
           \faTimes & \faCheck & \faTimes &33.10 &0.961
           \\
           \faTimes & \faTimes & \faCheck & 33.41 & 0.964
           \\
           \faCheck & \faCheck & \faTimes &33.13 &0.962
           \\
           \faCheck & \faCheck & \faCheck &33.55 &0.966
         \\
         \hline
    \end{tabular}}
\end{table}

\textbf{Effectiveness of each module.}
Table~\ref{tab:ablem} illustrates that the baseline model attains a PSNR of 32.83 dB. The inclusion of the Multi-input and Multi-output, SCAM, and HSFS modules leads to gains of 0.09 dB, 0.27 dB, and 0.58 dB, respectively, over the baseline model. When all modules are integrated simultaneously, our model achieves an improvement of 0.72 dB. This underscores the effectiveness of our proposed modules.

\textbf{Effectiveness of MSSFBlock.}
To assess the capability of MSSFBlock, we conducted several experimental designs incorporating different branch structures. As shown in Table~\ref{tab:msbabl}, compared to a single branch, the utilization of multiple branches in MSSFBlock leads to improved performance, indicating the effective extraction of contextual information. However, the accuracy saturates at branch 2, which is likely caused by an excessive amount of multi-scale information being captured. 

\begin{figure*}[htb] 
	\centering
	\includegraphics[width=1\textwidth]{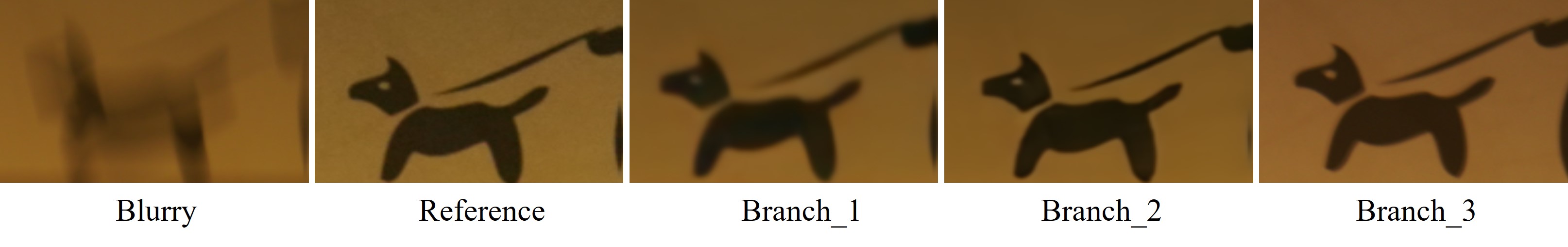}
	\caption{Visual results achieved by  different branch structures. Branch$\_i$ represent MSSFBlock has $i$ branches.}
	\label{fig:mssff}
\end{figure*}

To delve deeper into this observation, we present plots displaying the results obtained from various experimental designs. As depicted in Figure~\ref{fig:mssff}, it's clear that with an increasing number of branches, the overall structure of the results remains stable. However, there's a noticeable decline in accurately capturing fine details. This observation suggests that an excessive capture of multi-scale information leads to a disproportionate emphasis on contextual features over spatial details in the final feature representation.

\begin{table}
    \caption{The impact of MSFSBlock.}
    \label{tab:msbabl}
    \centering
    \resizebox{\linewidth}{!}{
    \begin{tabular}{ccccc}
    \hline
         Branch number&  Kernel size & $\#P$ &  PSNR & SSIM 
         \\
         \hline
        1 & 3x3 & 0.54M & 33.37&  0.9642
         \\ 
         2 & 3x3, 5x5 &0.57M &33.55&0.9664
         \\
         3 & 3x3, 5x5, 7x7&0.98M  & 33.54&0.9663
        \\
         \hline
    \end{tabular}}
\end{table}

\textbf{Effectiveness of MSFM.}
To assess the capability of MSFM, we compare our MSFM with several alternatives in Table~\ref{tab:msf}. It is clear to see that our MSFM achieves the final result. Obviously, it can be seen that our MSFM has achieved the final effect, which proves that our MSFM can effectively adaptively select the frequency with the largest amount of information for recovery.

\textbf{Design choices for skip connections.} 
As depicted in Table~\ref{tab:rwal}, we compare our SCAM with several skip connection alternatives.
In comparison to sum, concatenation, and CGB~\citep{tu2022maxim}, our SCAM achieves a higher value.
This indicates that our SCAM effectively merges features across various scales by discerning the information transmitted through skip connections. 

\begin{table}
    \caption{Alternatives for MSFM.}
    \label{tab:msf}
    \centering
   
    \begin{tabular}{ccccc}
    \hline
        Method &  Conv &  Wavelet & MDSF~\citep{FSNet} & MSFM
         \\
         \hline
        PSNR & 33.18 & 33.15 & 33.39 & 33.55
        \\
         \hline
    \end{tabular}
\end{table}

\begin{table}
    \caption{The impact of skip connections on the overall performance.}
    \label{tab:rwal}
    \centering
    \resizebox{\linewidth}{!}{
    \begin{tabular}{ccccc}
    \hline
         Modules&  Sum&  Concatenation & CGB~\citep{tu2022maxim} & SCAM
         \\
         \hline
         PSNR&  33.31  &33.33  &33.41&33.55
         \\ 
         \hline
         SSIM   & 0.9652& 0.9657 & 0.9661 &0.9663
         \\
         \hline
    \end{tabular}}
\end{table}

\subsection{Resource Efficient}
As deep learning models aim for higher accuracy, their complexity increases, demanding more resources. However, in certain contexts, deploying larger models may be impractical. Table~\ref{tab:computational} and Figure~\ref{fig:param} demonstrate that our LCDNet model achieves state-of-the-art performance while concurrently reducing computational costs. Specifically, on the Rain 100L dataset~\citep{Gopro}, we achieve the SOTA performance with up to 84.2\% cost reduction over the previous best approach, U$^2$former~\citep{u2former}. 
On the GoPro dataset~\citep{Gopro}, the inference speed of our method is faster. This highlights the efficiency of our method, showcasing superior performance along with resource effectiveness.
\begin{table}
    \centering
    \caption{The evaluation of model computational complexity.}
    \label{tab:computational}
\resizebox{\linewidth}{!}{
    \begin{tabular}{ccccc}
    \hline
         Method& Time(s) & MACs(G)  & PSNR & SSIM
         \\
         \hline\hline
         DBGAN~\citep{DBGAN}& 1.447 & 759 & 31.10 &0.942
         \\
         MPRNet~\citep{Zamir2021MPRNet} & 1.148 & 777 & 32.66 &0.959
         \\
         Restormer~\citep{Zamir2021Restormer} & 1.218 & 140 & 32.92 & 0.961
         \\
         Stripformer~\citep{Tsai2022Stripformer} &1.054 &170 & 33.08 &0.962
         \\
         IRNeXt~\citep{IRNeXt} &\underline{0.255} & 114 & 33.16 & 0.962
         \\
         SFNet~\citep{SFNet} &0.408 & 125 & 33.27 &\underline{0.963}
         \\
         FSNet~\citep{FSNet} &0.362 & \underline{111}& \underline{33.29} & \underline{0.963}
         \\
         \hline
         LCDNet(Ours) &\textbf{0.252} &\textbf{33} & \textbf{33.55} & \textbf{0.966}
         \\
         \hline
    \end{tabular}}

\end{table}

\section{Limitation}
Despite the advanced performance achieved by our method, our method treats all regions in the degraded image as degraded to the same degree. However, in real world scenarios, the degradation degree of different degraded regions is often inconsistent. In the future, we will consider differential processing for different degraded regions.

\section{Conclusion}
In this paper, we introduce LCDNet, a multi-scale design focused on learning both spatial details and contextualized information for image restoration. Our design includes HSFSBlock, which captures multi-scale information from the spatial domain and selectively chooses the most informative components for image restoration through channel attention in the frequency domain. Additionally, we develop SCAM to selectively determine the information propagating in skip connections, addressing noise introduced by conventional addition or concatenation. Extensive experiments demonstrate that LCDNet achieves state-of-the-art performance across four typical image restoration tasks.

\bibliographystyle{elsarticle-num}

\bibliography{refbib}

\end{document}